\documentclass[twoside,11pt]{article}

\usepackage[utf8]{inputenc}
\usepackage{amsmath}
\usepackage{amssymb}
\usepackage{physics}
\usepackage{mathtools}
\usepackage{amsthm}
\usepackage{tikz-cd} 
\usepackage[title]{appendix}
\usepackage{mdframed}
\usepackage{bbm}

\newmdtheoremenv{fthm}{Theorem}
\newmdtheoremenv{flem}{Lemma}
\newmdtheoremenv{fcor}{Corollary}
\newmdtheoremenv{fdef}{Definition}

\newcommand{\pard}[2]{\frac{\partial{#1}}{\partial{#2}}}
\newcommand{\R}{\mathbb{R}}
\newcommand{\set}[1]{\{ #1 \}}

\usepackage[abbrvbib, preprint]{jmlr2e}
\usepackage{soul}
\usepackage{hyperref}
\usepackage{algorithm}
\usepackage{bbm}

\ShortHeadings{Commutativity and Disentanglement from the Manifold Perspective}{Qiu}

\firstpageno{1}

\begin{document}

\title{Commutativity and Disentanglement from the Manifold Perspective}

\author{\name Frank Qiu \email frankqiu@berkeley.edu
        }
        
\maketitle

\begin{abstract}
In this paper, we interpret disentanglement as the discovery of local charts of the data manifold and trace how this definition naturally leads to an equivalent condition for disentanglement: commutativity between factors of variation. We study the impact of this manifold framework to two classes of problems: learning matrix exponential operators and compressing data-generating models. In each problem, the manifold perspective yields interesting results about the feasibility and fruitful approaches their solutions. We also link our manifold framework to two other common disentanglement paradigms: group theoretic and probabilistic approaches to disentanglement. In each case, we show how these frameworks can be merged with our manifold perspective. Importantly, we recover commutativity as a central property in both alternative frameworks, further highlighting its importance in disentanglement. 
\end{abstract}


\section{Introduction}
Disentanglement has many definitions but is broadly understood as the discovery of semantically meaningful factors that describe a dataset (\cite{bengio2013deep, bengio2014representation, higgins2018definition, burgess2018understanding}). On a seemingly unrelated note, there have been attempts to understand real-world data from the perspective of manifolds (\cite{LocallyLinearEmbed, manifoldLearningChapter, Tenenbaum_isomap, SparseManiTrans}); a prime example is the manifold hypothesis (\cite{TestManiHypFefferman}), which posits that high-dimensional data from the real world has a low-dimensional description. We argue that these two topics touch on the same idea, and  there is a natural interpretation of disentanglement as learning the data manifold's local charts. Indeed, this perspective has already been advanced several times (\cite{DeepLearnWorksManiHyp, zhou2021evaluating, LearnDisFactor}), linking manifold learning and disentanglement as the same task. In this paper, we give precise definitions to the informal concepts of disentanglement and factors of variation under this manifold framework, and we proceed to explore the implications of these definitions. Importantly, we show that commutativity between factors of variation is central to disentanglement. We demonstrate the utility of this framework applied to two problems - the learning of matrix exponential operators and the distillation of data-generating models. Finally, we conclude with a discussion of how our manifold framework relates to two other common disentanglement frameworks: group theoretic and probabilistic disentanglement.

\section{Disentanglement from the Manifold Perspective}

Most approaches to disentanglement involve learning a data-generating function whose arguments correspond to semantically meaningful variations in the data. That is, given a dataset $P$, one learns a generative function $f$ such that:
\[
\forall p \in P, \ \exists x_i \in X_i \quad \text{s.t.} \quad f(x_1,\cdots,x_n) \approx p
\]
and each $x_i$ captures something meaningful about the data. The above requirement might be unrealistic for real datasets where the data can vary in many distinct ways. For example, consider a collection of portraits of multiple people, taken under varying lighting conditions and poses. It would difficult to learn a global function $f(x_1,x_2)$ where $x_1$ controls lighting and $x_2$ controls pose, since there is so much variability in facial features between subjects. Instead, it might be more realistic to learn a local function $f_p$ that controls lighting and pose for each person $p$.

The above process of breaking up the generative function $f$ can be further extended: for many or all data points $p_i$ we learn a local generative function $f_{p_i}(x_1,\cdots,x_n)$ such that
\[
f_{p_i}(x_1,\cdots,x_n) \approx p'
\]
for all points $p'$ close to $p_i$. Hence, we pass from the strong requirement of a global function $f$ to the weaker requirement of a collection of local functions $f_{p_i}(x_1,\cdots,x_n)$. Intuitively, we can interpret the latent variables $x_i$ as the intrinsic coordinates of the local patches they describe, and we shall shortly formalize this notion.

\subsection{Defining Disentanglement}
The manifold hypothesis posits that high-dimensional data drawn from the real world lies on or near a low-dimensional manifold. More informally, high-dimensional data has a low-dimensional description. This description is given by the manifold's local charts, which map patches of the manifold to Euclidean coordinates. Returning to our disentangling function $f(x_1,\cdots,x_n)$, let us assume the latent variables $x_i$ are all real numbers. As previously discussed, we can interpret these disentangled latent variables as the data manifold's coordinates.

\begin{fdef}
The data lies on or near a smooth  manifold $M$, and the function $f(x_1,\cdots, x_n)$ \textbf{disentangles} the data if (locally) its latent variables $x_i$ correspond to the coordinates of some local chart of $M$. We call  $f(x_1,\cdots,x_n)$ a \textbf{disentangling function}.
\\
\end{fdef}

In practice, a learned generative map $f$ might not be a chart map itself but instead becomes a local chart after suitable restriction. For example, consider the positions of a clock's hour hand, which can be identified with the unit circle $\mathbb{S}^1$. We have the natural covering map $\pi$:

\[
\pi: \R \mapsto \mathbb{S}^1 \quad ; \quad t \mapsto e^{2 \pi i t} 
\]

This covering map describes the full set of hour hand positions but is not injective and hence cannot be a local chart. However, if we restrict $\pi$ to any suitably small interval $I$, then $\pi|_I$ is a local chart. Importantly, the latent variable of $\pi|_I$ coincides with the coordinate of a local chart. Note that our definition of disentanglement is a local one, and for the remainder of this paper we shall primarily concern ourselves with local properties of a disentangling function.

\subsection{Defining Factors of Variation}
Another prevalent concept in the disentanglement literature is that of a factor of variation. While its precise definition varies, there are a few properties that are generally agreed upon (\cite{bengio2014representation, LearnDisFactor}). Firstly, a factor of variation encodes a single semantically meaningful dimension on which the data can vary, and manipulating that dimension generates other valid data points. Going back the portraits example, the lighting condition of each portrait can be considered a single factor of variation. For a given portrait, we can slightly brighten or dim the lighting, and doing so would generate a new valid portrait. Secondly, a factor of variation has an implicit consistency condition, where sequential variations must be consistent with their total variation. Given a portrait $p_0$, suppose we dimmed its lighting condition by 10 lumens to generate a new portrait $p_1$, and then we dimmed the lighting condition of portrait $p_1$ by 10 lumens to generate another portrait $p_2$. If we dimmed the lighting of the original $p_0$ by 20 lumens, this should generate the same portrait $p_2$. Finally, different data points can be described by different factors of variation, and if a data point $p$ is unaffected by a change in a factor's value then it isn't described by that factor. In summary, we have the following basic properties of a factor of variation:

\begin{enumerate}
    \item \textbf{Feasibility:} At data point $p \in M$, varying the factor $F$ by small enough $t$ results in another data point:
    \[
    F(t,p) \in M
    \]
    \item \textbf{Consistency:} At data point $p$, varying factor $F$ by $t+s$ gives the same result as sequentially varying by $t$ and then $s$:
    \[
    F(t+s,p) = F(s,F(t,p))
    \]
    \item \textbf{Non-degeneracy:} If $p$ is described by the factor $F$, then varying the factor $F$ causes a change in the data. That is, there is no interval $I$ containing 0 such that $F(t,p)$ is constant for $t \in I$.
    
\end{enumerate}

The first two conditions characterize a factor of variation as a \textbf{flow function}, which is a function $F: D \rightarrow M$, where $D \subseteq \R \times P$, that satisfies the first two conditions. Intuitively, at each data point $p$ a factor's flow function describes a path through the dataset resulting from varying that factor. For example, given a portrait $p$, continuously changing its lighting condition generates a path of through the portrait dataset where each point on that path is a portrait $p_t$ that differs from $p$ in only its lighting condition. If we further assume that a factor's flow be smooth with respect to the data manifold's topology, we can formalize a factor of variation in the language of smooth manifolds:

\begin{fdef}
    A \textbf{factor of variation} $\boldsymbol{\theta}$ is a collection of smooth non-degenerate local flows on the data manifold $M$. If $p \in M$ is described by the factor $\theta$, there is a smooth non-degenerate local flow $\theta_p$ on a neighborhood of $p$ called a \textbf{local factor of variation}. \\
\end{fdef}

Note the locality condition in our definition. In a heterogeneous dataset, certain factors of variation might not be universal across the data. For example, not all mammals have tails, and tail length would be a factor of variation unique to a subset of mammals. The locality condition allows our definition to flexibly cover such cases. 

\subsubsection{Discrete Factors, parameterization, and Smoothness}
We assumed that each factor of variation is continuous, but in practice there are many examples of discrete factors. The subject's identity in our portrait dataset is one such example, where we have portraits of different people. However, discrete factors of variation can still be subsumed in our manifold framework. Continuous factors describe small local changes, and we will shortly establish that continuous factors correspond to variations in the data manifold's coordinates. On the other hand, discrete factors describe jumps in the data manifold. For example, in our portrait dataset, the collection of portraits of a single person constitute a cluster, and varying continuous factors like lighting conditions or camera angles keeps us within that cluster. However, varying the subject identity makes us jump from cluster to cluster, and in the language of manifolds we can imagine discrete factors as describing jumps between disjoint neighborhoods or even components of the data manifold. Hence, discrete factors describe a fundamentally different change in the data, and in keeping with our emphasis on local coordinates we shall consider only continuous factors.

We also assumed that each factor of variation was parametrized by a real number. Notably, in the group theory definition of disentanglement, a factor of variation may take values in some some group other than the reals. However, in Section \ref{sec:GroupBasedDis} we shall see that for a large class of naturally-occurring groups, group-valued factors can be equivalently described by a collection of real-valued factors.

Finally, we argue that our smoothness assumption is a mild one. Often, factors correspond to variations that are smooth with respect to the ambient Euclidean space the data lives in. If the data manifold $M$ is an immersed manifold of the ambient Euclidean space, then such factors are also smooth with respect to the smooth topology on $M$ because immersions are local embeddings. Since our low dimensional data manifold $M$ is assumed to live in a much higher dimensional Euclidean space, it seems reasonable to assume that $M$ is an immersed submanifold. Indeed, stronger results like the Whitney embedding theorems corroborate this.

\subsection{Factors of Variation and Disentanglement}
Having defined both disentanglement and factors of variation, we now discuss their relation within the smooth manifold framework. Recall that a disentangling function $f(x_1,\cdots,x_n)$ is a function whose latent variables locally correspond to coordinates of the data manifold. Within a local chart, suppose we varied the latent $x_i$ while holding all others fixed. This in fact generates a smooth local flow over the manifold, and each coordinate has an associated local flow over the data manifold. Conversely, given a data point $p$ described by factor $\theta$, we have a local flow $\theta_p$ around $p$. It turns out that there exists a local chart such that $\theta_p$ corresponds to a coordinate of that chart.

\begin{fthm}[Local Factors of a Disentangling Function]\label{Thm:FacDisRel}
    Suppose our data lies on a smooth manifold $M$, and let $f: \R^n \rightarrow M$ be a disentangling function.
    \begin{enumerate}
        \item For open $U \subseteq \R^n$, if $f|_U$ is a local chart of $M$ then each latent $x_i$ of $f|_U$ corresponds to a unique local factor of variation $\theta_i$.
        \item Every factor of variation can be locally realized as the latent variable of a disentangling function.\\
    \end{enumerate}
\end{fthm}

\subsection{Semantic Meaning}
In our definitions of disentanglement and factors of variation, we made no mention of semantic meaningfulness. This was intentional, as semantic meaning is a subjective quality left to the interpretation of the observer. Moreover, there are often multiple ways to meaningfully describe the same data. In language of manifolds, there are multiple meaningful coordinate systems for the data manifold, and the choice of a particular coordinate system is arbitrary. Therefore, we focus on structural properties common to all coordinate systems, and the issue of choosing a particular one is left untouched.

\section{Commutativity and Disentanglement}
Theorem \ref{Thm:FacDisRel} of the previous section shows that locally every disentangling function gives rise to a collection of distinct factors. What about the converse: can a set of local factors be realized as the latent variables of a disentangling function? In this section, we establish commutativity between factors as a necessary and sufficient condition for their joint disentanglement. This commutativity criterion serves as a guiding principle for designing disentangling algorithms and suggests that commutativity is a useful "inductive bias" - properties that are \textit{a priori} baked into a learning algorithm to encourage the emergence of good representations (\cite{GoyalInducBias}). In this vein, in Section \ref{sec:MatrixExpo} we demonstrate the utility of enforcing commutativity in the problem of learning matrix exponentials that describe the data.

\subsection{Local Dynamics of a Factor}
Before we get to the main result of this section, we first establish a relevant technical notion. Recall  that locally, a factor of variation is a smooth local flow $\theta(t,p)$ on some patch of the data manifold. Working in local coordinates, consider the function $X(p) = \pdv{}{t}|_{t=0} \theta(t,p)$ that assigns to each point $p$ the instantaneous change dictated by $\theta$. $X(p)$ captures the local dynamics of factor $\theta$, and $X(p)$ is the manifold analogue of a vector field over $\R^m$ encountered in vector calculus. This function $X(p)$ is in fact independent of our choice of coordinates, and for every local factor $\theta$ we have the associated local vector field $X_\theta$. Furthermore, just like how Euclidean vector fields form a real vector space, (local) vector fields over a manifold also have a vector space structure.  Hence, we make the following definitions:
\begin{fdef}
    Given a local factor of variation $\theta$, we call its associated local vector field $X(p) \equiv \pdv{}{t}|_{t=0} \theta(t,p)$ the \textbf{local dynamics of factor} $\boldsymbol{\theta}$. Two vector fields are \textbf{linearly independent} if they are pointwise linearly independent.\\
\end{fdef}

\subsection{The Commutativity Criterion}
We first define what it means for a collection of factors to commute. Returning to our dataset of portraits, each portrait is an image on a two-dimensional pixel grid. Suppose we generated new images by manipulating the spatial orientation of the portrait subject by two operations: rotating the subject about the center of the pixel grid and translating (shifting) the subject. It is well known that these two operations are not interchangeable: the resulting image depends on the order of their application. On the other hand, changing the lighting condition or shifting the subject results in the same image regardless of the order of their operation. The former is an example of non-commuting factors, while the latter is an example of commuting factors. With this definition of commutativity, we now state the central result of this paper.

\begin{fdef}
    Given a set of local factors $\set{\theta_i}$, they \textbf{commute} if:
    \[
    \theta_i (s, \theta_j(t,p)) = \theta_j (t, \theta_i(s,p))
    \]
    for all pairs $i,j$ and for all $s,t,p$ that lie in the flows' domain. In other words, two factors commute if the end result does not depend on their order of application.\\
\end{fdef}

\begin{fthm}[The Commutativity Criterion]\label{Thm:CommCrit}
    Let $\theta_1,\cdots,\theta_k$ be a set of local factors whose local dynamics $X_i$ are linearly independent. The factors can be locally disentangled - ie. they are the coordinates of a local chart - if and only if they commute.\\
\end{fthm}

The condition of linear independence is a technical condition that enforces the factors to be distinct. For example, any factor commutes with itself, and linear independence ensures we do not have copies of the same factor. Theorem \ref{Thm:CommCrit} shows that commutativity is a central property of disentanglement: the coordinates of a local chart define commuting factors, and a set of factors can be realized as the coordinates of a local chart only if they commute.  Note that Theorem \ref{Thm:CommCrit} does not require that the number of factors equal to the dimension of the data manifold: we can apply the commutativity criterion to any set of factors, even if they only partially describe the data.

\subsection{Example: Grid-Centric Rotations and Translations}
The commutativity criterion implies that it is impossible for a system of non-commuting factors to be jointly disentangled. Consider again the example of grid-centric rotation and translation applied to the portrait dataset. Each operation is a factor of variation, and their disentanglement amounts to a local chart $\phi$:
\[
\phi(x_1,x_2,x_3) = p
\]
such that $x_1$ corresponds to grid-centric rotation and $x_2,x_3$ correspond to translations. However, as previously mentioned grid-centric rotation does not commute with translation, so no such $\phi$ can exist by Theorem \ref{Thm:CommCrit}. It is impossible for any disentangling function to represent these operations. 

Now suppose we took a single portrait and generated a new dataset by applying grid-centric rotations and translations to the portrait subject. Is it now possible to learn a function that disentangles the two operations? Again, Theorem \ref{Thm:CommCrit} asserts that this is impossible. However, we explicitly generated the data through these two operations, so what went wrong?  For example, we can certainly define a function that generates the entire dataset by sequentially applying translation and rotation to the original portrait $p_0$:
\[
g(x_1,x_2,x_3) = R(x_1) T(x_2,x_3) p_0
\]
where $R(x_1)$ is rotation and $T(x_2,x_3)$ is horizontal and vertical translation respectively. \textbf{The issue is that the latent variables do not correspond to the original data-generating operations}. Suppose we fixed $x_1 = \frac{\pi}{4}$, so we always apply a fixed rotation after translation. Then, varying $x_2$ would not translate the portrait subject horizontally but instead translates them along an axis rotated $\frac{\pi}{4}$ from the $x$-axis. Similarly, varying $x_3$ translates the subject on an axis rotated $\frac{\pi}{4}$ from the $y$-axis. The factors defined by $x_2$ and $x_3$ do not correspond to the original data-generating factors of horizontal and vertical translation!

Now say we flipped the order of application, so we first apply rotation and then translation:
\[
h(x_1,x_2,x_3) = T(x_2,x_3)R(x_1) p_0
\]
Suppose the subject of the original portrait $p_0$ was centered such that their nose was in the center of the pixel grid, and we fix the translation parameters $(x_2,x_3)$ to $(3,3)$ so the subject is displaced relative to the grid's center. Now if we vary the rotation parameter $x_1$, this would result in the subject rotating about their nose displaced $(3,3)$ units relative to the grid center. In other words, the factor $x_1$ captures nose-centric rotation rather than grid-centric rotation. Again, we have a factor that does not correspond to the original data-generating factors.

\subsection{Local Dynamics of Commuting Factors}
In fact, we can more precisely analyze the previous example to demonstrate an interesting equivalent characterization of commutativity. Recall that we generated the data by applying grid-centric rotation and translation to the subject of a single portrait $p_0$. Let us assume that images have infinite dimension and infinite resolution, so our data can be identified with the manifold $\mathbb{S}^1 \times \R^2$. Around $p_0$, we have the local coordinates $(x_1,x_2,x_3)$ that track orientation and position relative to $p_0$ respectively, so $p_0$ has coordinates $(0,0,0)$. We again consider the data-generating function $g(x_1,x_2,x_3) = R_{x_1} T(x_2,x_3) p_0$, which in local coordinates has the form:
\[
g: (x_1,x_2,x_3) \mapsto (x_1, R_{x_1} [x_2,x_3])
\]
where $R_{x_1}$ is the $2 \times 2$ rotation matrix of angle $x_1$.
Computing the Jacobian of $g$ in this coordinate basis, we see that the associated local coordinate dynamics of $g$ are:
\begin{align*}
    X_1 &= \pard{}{x_1} + (-x_2 \sin x_1 - x_3 \cos x_1)\pard{}{x_2} + (x_2 \cos x_1 - x_3 \sin x_1) \pard{}{x_3}\\
    X_2 &= \cos x_1 \pard{}{x_2} + \sin x_1 \pard{}{x_3}\\
    X_3 &= -\sin x_1 \pard{}{x_2} + \cos x_1 \pard{}{x_3}
\end{align*}
Notably, the local dynamics of all three factors involve the angle of rotation $x_1$, and their dynamics depend on each other! This explains why grid-centric rotation and translation cannot be disentangled.

The coupling between the local dynamics $X_i$ of a pair of factors $\theta_i$ can be computed through their Lie derivative, denoted $L_{X_1} X_2$. Roughly, $L_{X_1} X_2$ computes the change in the local dynamics of factor $\theta_2$ when varying $\theta_1$. Hence, the local dynamics of the factors are decoupled if they do not depend on other factors, and we formalize this concept.

\begin{fdef}
    For local factors $\set{\theta_i}$ with local dynamics $\set{X_i}$, we say their \textbf{local dynamics are decoupled} if:
    \[
    L_{X_i} X_j = L_{X_j} X_i = 0 \qquad \forall i \neq j
    \]
    That is, the local dynamics of each factor do not depend on the values of other factors.\\
\end{fdef}

\begin{fthm}[Commutativity and Local Dynamics] \label{thm:flowVecEequiv}
    A set of local factors commute if and only if their local dynamics are decoupled.\\
\end{fthm}
Theorem \ref{thm:flowVecEequiv} gives us another way of interpreting why commutativity between factors of variation is so important: it captures the property that a factor of variation affects the data in the same way regardless of the other factors' values. This captures the desiderata that factors of variation encode separate qualities of the data (\cite{bengio2014representation}), which by Theorems \ref{Thm:CommCrit} and \ref{thm:flowVecEequiv} is equivalent to commutativity between factors.

\section{Application to Matrix Exponential Operators}\label{sec:MatrixExpo}
In this section, we apply the framework developed so far to algorithms that learn data-generating operators through the matrix exponential (\cite{cohenLieGroup, Miao2007_learnLG, sohldickstein2017unsupervised, xiaoContFlow, chau2020disentangling, XuLieGruopLearn}). While these algorithms might vary in guise and name, their core approach is the same: learning a suitable dictionary of matrices so that their matrix exponential results in a broad of set of data-generating operators. By applying these operators to a data point, one generates new data points that differ from the original in interesting, meaningful ways. We shall see that enforcing commutativity has both theoretical and practical benefits to this class of algorithms. In a later section, we also discuss their relation to the group theoretic definitions of disentanglement.

\subsection{Problems with the Matrix Exponential}
Computing a matrix exponential is relatively expensive (\cite{MolerMatExpCost}), and a common computational shortcut  (\cite{ cohenLieGroup, sohldickstein2017unsupervised}) is to assume the generators $A_i$ are jointly diagonalizable: $A_i = PD_iP^{-1}$ for some diagonal matrix $D_i$ with common matrix $P$. For such matrices, the matrix exponential is easy to compute and takes the following form:
\[
\exp(\sum \alpha_i A_i) = \exp(\sum \alpha_i P D_i P^{-1}) = P\exp(\sum \alpha_i D_i)P^{-1} 
\]
where $\exp(\sum \alpha_i D_i)$ is a diagonal matrix whose entries are the entry-wise exponential of the diagonal matrix $\sum \alpha_i D_i$. However, the assumption of joint diagonalization is often not a natural one and is instead invoked as a shortcut.

Another problem when working with sums of matrix generators is that the property
\[
e^{sA + tB} = e^{sA}e^{tB}
\]
generally holds for all $t,s$ only if $A$ and $B$ are commuting matrices (this directly follows from the Baker-Campbell-Hausdorff formula). This has an important consequence for interpreting the impact of each dictionary element's coefficient. It is well-known that the matrix exponential has the property: $\pdv{}{t}|_{t=0} \exp{tA} = A$. However, this property fails to extend to sums of non-commuting matrices: $\pdv{}{\alpha_i}|_{\alpha_i=0} \exp(\sum \alpha_i A_i) \neq A_i$. In particular, this means that the data change induced from varying $\alpha_i$ does not equal the local flow $\theta_i(\alpha_i,p) = \exp(\alpha_i A_i)p$. This makes the effect of $\alpha_i$ hard to interpret, since its dynamics might be considerably more complicated than the local flow of $A_i$.

\subsection{Commutativity and the Matrix Exponential}
Let us examine the task of learning matrix exponentials from the manifold perspective. Applying the matrix exponential to a point $p$, we have the local data-generating function:
\[
\phi(\alpha_1,\cdots,\alpha_n) = e^{\sum \alpha_i A_i} p
\]
Again, we associate each dictionary element $A_i$ to its canonical factor $\theta_i(t,p) = \exp(t A_i) p$. Under what conditions do the coordinates of $\phi$ reflect these dictionary factors? We apply the commutativity criterion of Theorem \ref{Thm:CommCrit} to establish a simplified form of the matrix exponential.

\begin{fthm}[Commutativity and the Matrix Exponential]\label{thm:CommMatExp}
    Consider a data-generating operator of the form $\exp(\sum \alpha_i A_i)$ from a dictionary of linearly independent square matrices $A_i$. At each point $p$, the coordinates of the data-generating function
    \[
    f_p(\alpha_1,\cdots,\alpha_n) = \exp(\sum \alpha_i A_i) p
    \]
    correspond to the dictionary factors $\theta_i(t,p) = \exp(t A_i)p$ if and only if the dictionary elements $A_i$ commute as matrices. Furthermore, the $A_i$ can be simultaneously diagonalized (possibly over $\mathbb{C}$) and the matrix exponential splits into a product of the dictionary factors.
\end{fthm}

Commutativity of the dictionary elements $A_i$ is precisely the condition that allows them to be realized as the factors of a local chart. Furthermore, it justifies the computational shortcut of assuming joint diagonalization and makes it easy to interpret the coefficients of the model. Viewing the data-generating operator as a disentangling function, the manifold disentanglement framework tells us to restrict our attention to dictionaries of commuting matrix generators, which are both easy to work with and interpret.

\section{Distilling and Disentangling Data-Generating Models}
In the previous section, we examined disentanglement for a specific type of data-generating model: the matrix exponential. In this section, we show how any smooth data-generating model can be locally distilled and disentangled at points where it is full rank. This implies a possibility result for distilling complex data-generating models into a more compact one,  and this compact representation is (almost) the smallest one that still preserves the descriptive power of the original model.

We first define the notion of a data-generating model's local expressivity. Intuitively, a model's local expressivity captures the number of directions one can move the data manifold $M$ by small variations in the model's latent variables. In coordinates, this notion is naturally captured by the rank of model's Jacobian, and at such points a joint distillation and disentanglement procedure is possible.

\begin{fdef}
    Let $f: \R^m \rightarrow M$ be a smooth data-generating function describing the $n$-dimensional data manifold $M$. For $x \in \R^m$, the \textbf{local rank of} $\boldsymbol{f}$ \textbf{at} $\boldsymbol{x}$ is the rank of its differential $\partial f_x$, which in coordinates is the Jacobian of $f$. We say $f$ is \textbf{full rank at} $\boldsymbol{x}$ if $\partial f_x$ is full rank.\\
\end{fdef}

\clearpage
\begin{fthm}[Distillation and Disentanglement of Models]\label{thm:DistDisenModel}
    Let $f: \R^m \rightarrow M$ be a smooth data-generating function describing the $n$-dimensional data manifold $M$. If $f$ is full rank at $x \in \R^m$, then there exists a smooth reparametrization $\psi$ of an open set $V$ containing $x$ and a local chart containing $f(x) \in M$ with the following property:
    \begin{enumerate}
        \item If the rank of $f$ is $m$, in coordinates $f \circ \psi^{-1}$ takes the form:
        \[
        f \circ \psi^{-1}: (x_1,\cdots,x_m) \mapsto (x_1,\cdots, x_m,0,\cdots,0)
        \]
        \item If the rank of $f$ is $n$, in coordinates $f \circ \psi^{-1}$ takes the form:
        \[
        f \circ \psi^{-1}: (x_1, \cdots, x_n,\cdots,x_m) \mapsto (x_1,\cdots,x_n)
        \]
    \end{enumerate}
\end{fthm}

\begin{fcor}[Distillation of Overcomplete Models] \label{cor: DisMaxModels}
    Let $f: \R^m \rightarrow M$ be a smooth data-generating function describing the $n$-dimensional data manifold $M$, with $m \geq n$. If $f$ is surjective and full-rank everywhere, then:
    \begin{enumerate}
        \item $f$ can be locally distilled and disentangled using $n$ parameters at every point in its latent space.
        \item $f$ can be globally distilled into a function $f^*$ defined on an subset of $\R^{n+1}$ that is surjective and full-rank with respect to the first $n$ latent variables.
    \end{enumerate}
    \end{fcor}

Theorem \ref{thm:DistDisenModel} shows that there exists a reparameterization $\psi$ of the latent space of a model such that a subset of the reparametrized variables capture all of the possible variation of $f$. In this sense, the distillation procedure is similar in spirit to dimensionality reduction techniques, and it can be seen as their analogue for data-generating models. Moreover, we see that these reparameterized variables $\psi(x)$ correspond to manifold coordinates, meaning that the data-generating function now also locally disentangles the data. 

Corollary \ref{cor: DisMaxModels} captures the common situation of a complex, overcomplete model that fully describes the data. It not only establishes that such overcomplete models can be locally distilled and disentangled, one can also stitch these local distillations into a global one using at most one extra parameter. This alternative model $f^*$ is just as expressive as $f$ while having the (almost) minimal number of latent variables, and its latent space is essentially a reparametrized subset of the original latent space of $f$. Hence, Corollary \ref{cor: DisMaxModels} gives a possibility result that one can always distill an overcomplete model into a more compact one. 


\section{Connection to Group Theoretic Disentanglement}\label{sec:GroupBasedDis}
In this section, we highlight the connections between our manifold framework and group theoretic approaches to disentanglement.The group theoretic framework equips the data $M$ with a group of symmetries $G$ that act on it, and one seeks to factorize this action into a product of group factors $G = G_1 \times \cdots \times G_n$ acting on $M$ (\cite{higgins2018definition, WangGroupDisentangle}). Each individual group factor $G_i$ controls its own set of data properties, and they are the group theoretic versions of factors of variation. Under this framework, disentanglement amounts to learning a factorized group action on the data.

In following subsections, we will show that commutativity is an implicit assumption in group factorization. In fact, as is the case in our manifold framework, commutativity is a necessary and sufficient condition for group theoretic disentanglement. Furthermore, many groups of physical significance are Lie groups, and we will show how the group theoretic framework can be recast in the manifold framework when working with Lie groups.

\subsection{Commutativity and Factorized Group Actions}
As previously mentioned, disentanglement in the group theoretic framework amounts to factorizing the group of data symmetries $G$ into a product of subgroups $G = G_1 \times \cdots \times G_n$, where each group factor $G_i$ represents an independent set of data properties. For example, in our portrait dataset, the portraits of a subject might be governed by their orientation and position relative to the camera as well as the lighting condition, and the corresponding group of symmetries would be $G = SO(3) \times \R \times \R^2$. Disentanglement would entail recovery the factorized group structure and its corresponding action on the data.

Interestingly, the product structure of the factorized group implies commutativity between the group factors $G_i$. Each factor $G_i$ corresponds to the subgroup $(e,\cdots,G_i,\cdots,e) \subseteq G$, and these subgroups naturally must commute. For example, multiplication in the product group $G = G_1 \times G_2$ has the following property:
\[
(g_1,e) * (e,g_2) = (g_1,g_2) = (g_2,e) * (e,g_1) \qquad \forall g_1 \in G_1, \forall g_2 \in G_2
\]
This holds for any product group, even if the factors themselves are not abelian groups.

Conversely, suppose we have a set of factors of variation, which in the group theoretic framework are groups $G_i$ acting on the data $M$. A group action is equivalently a group homomorphism from $G_i$ into the automorphism group of $M$, and our set of factors is equivalent to a set of group homomorphisms $f_i: G_i \rightarrow Aut(M)$. By universality, these homomorphisms must filter through the group coproduct $\coprod G_i$:
\[
\begin{tikzcd}
G_i \ar[r] \ar[dr, "f_i"]  & \coprod G_i \ar[d, "\coprod f_i"]\\
& Aut(M)
\end{tikzcd}
\]
However, the group coproduct is the free product, not the product group as required under the group theoretic framework. This is because the factors $G_i$ might not commute, and the construction of the induced map $\coprod f_i$ is order-dependent. Intuitively, this is incompatible with disentanglement, since each $G_i$ governs a separate set of data symmetries: the order should not matter when changing different types of symmetries. Again, commutativity seems to capture some instrinsic quality of separateness between factors, and in fact there is a corresponding commutativity criterion for group theoretic disentanglement.

\clearpage
\begin{fthm}[Commutativity Criterion for Group Theoretic Disentanglement] \label{Thm:GroupCommCrit}
    Consider a set of group theoretic factors of variation, which are groups $G_i$ acting on the data $M$. They are the factors of a product group action $f: G = G_1 \times \cdots \times G_n \rightarrow Aut(M)$ if and only if their actions commute.
\end{fthm}

\subsection{Lie Groups and Manifold Disentanglement}
Lie groups, groups that are also smooth manifolds, serve as a bridge between the group theoretic framework and our manifold framework. In the case of a Lie group $H$, a smooth action of $H$ on the data $M$ is a group action that is also smooth. In our manifold framework, factors of variation are in fact also local smooth actions of $\R$ on the data manifold $M$, at least restricted to an open interval. Given a smooth action of $H$ on $M$, at each $p \in M$ the data manifold  locally inherits a (partial) coordinate chart from $H$ via its orbit map $f_p(h) = h \cdot p$. If $H$ is a product group, then $M$ inherits a coordinate chart that is a product of each factor $H_i$'s coordinates.

\begin{fthm}[Group-Manifold Disentanglement Correspondence]\label{thm:GroupManiCorres}
    For a product Lie group $H = H_1 \times \cdots \times H_k$ acting smoothly on the data manifold $M$, at every point $p \in M$ there exist coordinate charts containing $e \in H$ and $p \in M$ respectively such that the orbit map $f_p$ has the following coordinate form:
    \[
    f_p : \prod_{i=1}^k (x_{i1},\cdots,x_{in_i}) \mapsto \prod_{i=1}^k (x_{i1},\cdots,x_{ij_i})
    \]
    where $n_i = dim(H_i)$ and $j_i \leq n_i$. In other words, at every $p \in M$ there exists a factorized set of coordinates around $p$ such that the $i^{th}$ factor is induced from a subset of the coordinates of $H_i$.  The number of dimensions $j_i$ described by each $H_i$ is constant within the same orbit.
\end{fthm}
This results links disentanglement between the group theoretic and manifold frameworks. The smooth action of any factorized Lie group $H$ induces a correspondingly factorized coordinate chart at each point of $M$. Each subgroup $H_i$ acts on a different set of coordinates of $M$, and so we have groupings of local coordinates of the data that are controlled by distinct symmetry groups $H_i$.

In fact, a Lie group $H$ that acts transitively on the data manifold, meaning we can transform any data point to another via its action, is a model of $M$ itself. For example, consider the manifold of a subject's portraits. As previously mentioned, the orientation, position, and lighting of the portraits all naturally correspond to the Lie groups $SO(3)$, $ \R^2$, and $\R$ respectively. The joint action of these groups allows us to switch between any two of their portraits, and fixing a reference portrait $p_0$ we can identify each portrait $p$ with a unique group element $g_p$ such that $p = g_p \cdot p_0$. In this sense, the subject's portrait manifold is equivalent to the group of symmetries acting on it. Formally, such spaces are called homogeneous spaces and are essentially quotient manifolds of the symmetry Lie group.

\begin{fthm}(Homogeneous Spaces and Disentanglement)\label{thm:HomSpace}
    Suppose Lie group $H$ acts transitively on data manifold $M$, meaning the orbit map is surjective for every $p \in M$. Then $M$ is diffeomorphic to a quotient manifold of $H$. If $H$ also acts freely, meaning the orbit map is injective for every $p \in M$, then $M$ is diffeomorphic to $H$ itself.
\end{fthm}

\subsection{Revisiting Matrix Exponentials}
As a brief aside, we return to the problem of learning matrix exponentials of Section \ref{sec:MatrixExpo}. Given a dictionary of matrices $A_i$, the set of matrix exponentials of the form $\exp(\sum \alpha_i A_i)$ forms a Lie group and in many cases is an embedded Lie subgroup of the general linear group. Each dictionary element $A_i$ is a member of the Lie algebra of $H$, which intuitively gives a linear picture of $H$ around the identity element. The matrix exponential provides a local diffeomorphism between the Lie algebra and its Lie group,  and so the dictionary coefficients $\alpha_i$ in fact correspond to a local chart of $H$ provided the $A_i$ are linearly independent. By Theorem $\ref{thm:GroupManiCorres}$, the data manifold $M$ in turn inherits (some of) these coordinates, so the Lie group coordinates $\alpha_i$ also serve as manifold coordinates when applying these operators.

Each element of the Lie algebra defines a unique global flow over $H$, and any commuting, spanning set of dictionary elements $A_i$ in fact defines a global chart over $H$. Such a commuting Lie algebra basis implies that the identity component of $H$, which is a Lie subgroup, is an abelian group. Thus, working with a commuting dictionary means locally modeling the action of an abelian Lie group, which are all of the form $\R^n \times \mathbb{T}^m$ where $\mathbb{T}^m$ is the $m$-dimensional torus. Such groups are particularly simple to understand: for example, the representation theory of such groups is well-known and easy to compute (\cite{cohenLieGroup, chau2020disentangling}). This reinforces the benefits of commutativity in simplifying the matrix exponential problem. 

\section{Connection to Probabilistic Disentanglement}
In this section, we highlight the connections between probabilistic definitions of disentanglement and our manifold framework. Broadly, under the probabilistic framework one hopes to learn a factorized latent representation of the data $Z = (Z_1,\cdots,Z_k)$ such that the factors $Z_i$ are independent (\cite{HigginsbetaVAE, KimDisByFact, burgess2018understanding}). This latent representation might be the latent code of an autoencoder (\cite{KimDisByFact}) , the sources of a signal mixture  (\cite{ICA_Hyvarinen}), or even the principal axes of the data distribution as in principal component analysis. However, in all cases the factorization into independent $Z_i$ is central, and these factors $Z_i$ are probabilistic versions of factors of variation. We will detail the effects of commutativity within this probabilistic framework and show how it can be merged with our manifold framework.

\subsection{Probability Measures in Coordinates}
Under the probabilistic framework, the latent representation $Z = (Z_1,\cdots,Z_k)$ gives rise to the data $M$ via a generative function $p(M|Z)$. From the manifold perspective, we can naturally interpret the latent representation $Z$ as local coordinates of $M$ and the generative function $p(M|Z)$ as its (possibly noisy) inverse chart map. Assuming $Z$ is a set of full coordinates describing the manifold patch $U \subseteq M$, any probability measure on $Z$ induces a unique local measure on $U$ and vice versa. Now, if we have another coordinate system for $U$, how are the two coordinate measures related? It turns out that these coordinate measures must be consistent with one another and are essentially the same measure under different parameterizations.

\begin{flem}[Consistency of Local Coordinate Measures]\label{Lem: CoordConsistent}
    For the data manifold $M$, consider any measure $\mu$ on $M$ with respect to its Borel $\sigma$-algebra. For any open $U \subseteq M$, let $\phi_1,\phi_2: U \rightarrow \R^n$ be two local charts. Then,
    \begin{enumerate}
        \item For $i \in \set{1,2}$, $\mu|_U$ induces the unique coordinate measure $\mu_i$ on $\phi_i(U)$ via the pushforward map: $\mu_i = \phi_i^* (\mu|_U)$.
        
        \item Any two coordinate measures are related to each other via their transition map $\psi$:
        \[
        \mu_2 = \psi^*(\mu_1)
        \]
        \item All three $\phi_1,\phi_2,\psi$ are invertible measurable maps.
    \end{enumerate}
\end{flem}

\subsection{On the Impossibility of Probabilistic Disentanglement}
We now tackle an impossibility result stating that probabilistic disentanglement is impossible in a general setting (\cite{Locatello_disentanglement}). \textbf{For the rest of this section, we will assume that the data $X$ lies near a manifold $M$ rather than exactly on it, reflected in our notation.} The distinction of the data lying on or near the manifold did not matter in previous sections, but we adopt it now for clarity and convenience. 

The impossibility result states that given a latent measure $P(Z)$ and likelihood $P(X|Z)$, there exist infinitely many $\psi: Z \rightarrow Z$ such that $\psi(Z) \stackrel{d}{=} Z$ and the Jacobian $\partial \psi$ has non-zero entries.  The alternate latent variables $\Tilde{Z} = \psi(Z)$ are entangled with respect to $Z$ since each entry depends on all of $Z$, yet they induce the same data distribution $P(X)$. Therefore, the two parameterizations $Z$ and $\Tilde{Z}$ are indistinguishable when inferring their values from the data, and the recovery of $Z$ is ill-posed.

Viewing this situation from the manifold perspective, the two latent variables $Z$ and $\Tilde{Z}$ are different local coordinate systems of the manifold $M$. By the consistency requirements of Lemma \ref{Lem: CoordConsistent}, their measures must be pushforwards of each other through their transition map $\psi$, and they induce the same local measure on the manifold. Now, the crux of the issue lies in the form of the likelihood $P(X|Z)$. Consider the following data generating process:
\[
Z \rightarrow M \rightarrow M + \epsilon = X
\]
where $\epsilon$ is Gaussian noise. In other words, conditioned on a particular point $p \in M$, the data $X$ is normally distributed about $p$. If $\phi$ is the chart map of $Z$, then the likelihood is $P(X|Z) = \mathcal{N}(\phi^{-1}(Z), I)$. A similar situation holds for $\Tilde{Z}$ and its chart map $\Tilde{\phi}$. Importantly, we see that if a latent point $z$ maps to different points on the manifold under each chart, then we have $P(X|Z = z) \neq P(X| \Tilde{Z} = z)$ since the two Gaussians have different means. Even if $Z$ and $\Tilde{Z}$ have the same distribution, they can be still distinguished via their likelihoods! We can extend the conclusion of this particular example to a more general result.

\begin{fthm}[Likelihoods Distinguish Local Coordinates]\label{thm:DistinguishLocalCoords}
    Let $Z$ and $\Tilde{Z}$ be distinct (partial) local coordinates of the same open $U \subseteq M$. For data $X$ lying on or near the manifold $M$, suppose the family of likelihoods $\set{P(X|p)|p \in M}$ is identifiable in $M$. Then, there exists an open set $V$ such that:
    \[
    P(X| Z = v) \neq P(X|\Tilde{Z} = v) \qquad \forall v \in V
    \]
\end{fthm}

When $Z$ and $\Tilde{Z}$ have non-negative densities with respect to the Lebesgue measure, Theorem \ref{thm:DistinguishLocalCoords} implies that their likelihoods differ on a set of positive probability. Hence, while it might not be possible to distinguish between two local coordinates based on their distributions, one can distinguish between them using their likelihoods. Moreover, note that we can still apply Theorem \ref{thm:DistinguishLocalCoords} when we have only a partial description of the manifold, so even an incomplete set of local coordinates can be distinguished.

Indeed, in our manifold framework, the impossibility result essentially states that given a local coordinate system, there are infinitely many other coordinate systems whose transition maps preserve its coordinate measure. The consistency condition of Lemma \ref{Lem: CoordConsistent} ensures that any two coordinate measures induce the same measure on $M$ and hence on the data $X$. Therefore, it is impossible to infer the values of a particular set of coordinates from the data,  since there are are infinitely many coordinate charts with the same measure that describe a manifold patch. Instead, one must choose a particular coordinate system independently of the data by fixing the likelihood and hence local chart map, making the problem of recovering its coordinates become well-posed. \textbf{In light of this, the task of finding a factorized latent representation $Z$ requires the recovery of a local chart map, and the goal of probabilistic disentanglement aligns with the goal of manifold disentanglement.}

\subsection{Commutativity and Probabilistic Disentanglement}
Here, we establish an analogous commutativity criterion for probabilistic disentanglement in a Bayesian setting. Recall that factors of variation in our manifold framework are local flows over the data manifold, and we will use the shorthand $\theta_{t_i} p \equiv \theta_i(t_i,p)$. Given a set of local factors $\theta_1,\cdots,\theta_k$ at $p \in M$, we have a local data-generating function $f: T \rightarrow M$ defined on $T \subseteq \R^k$:
\[
f(t_1,\cdots,t_k) = \theta_{t_1} \circ \dots \circ \theta_{t_k} p
\]
Any probability measure over the factor values $T = (T_1,\cdots,T_k)$ naturally induces a corresponding local measure on $M$ and hence on $X$. However, if the factors $\theta_i$ do not commute, then the data-generating function depends on their order of application and knowing just the factor values $T$ does not specify a location on the manifold. Treating the order of application $\sigma$ as a random variable, our likelihood then becomes a mixture:
\[
P(X|T = T_1,\cdots,T_k) = \sum_{\sigma \in S_k} P(\sigma) P(X|T;\sigma)
\]
where $P(X|T;\sigma) = P(X| f(t_{\sigma(1)},\cdots,t_{\sigma(k)})$. Non-commutativity introduces ambiguity in the manifold location, expressed as a mixture of different likelihoods. For example, again suppose the data $X$ is conditionally Gaussian about the manifold $M$. Then, non-commutativity of the factors causes $P(X|T)$ to be a mixture of Gaussians, and the mixture becomes only becomes a single Gaussian when the factors commute. We generalize this example to the following result.

\begin{fthm}[Commutativity and Probabilistic Disentanglement]\label{thm:CommuCritProb}
    Let $\theta_1,\cdots,\theta_k$ be a set of local factors at $p \in M$, and let $T = (T_1,\cdots,T_k)$ be their joint time domain at $p$. Suppose the family of likelihoods $\set{P(X|m)| m \in M}$ is identifiable in $M$ and affinely independent. The likelihood of $X$ given $T$ takes the form:
    \[
    P(X|T) = \sum_{\sigma \in S_k} P(\sigma) P(X|T;\sigma) = \sum_{\sigma \in S_k} P(\sigma) P(X|f_\sigma(T))
    \]
    The likelihood $P(X|T)$ is not a mixture if and only if the factors commute.
\end{fthm}

In a probabilistic setting, commutativity between the factors means that the manifold location is fully specified from the flow parameters $T$ alone, as the order of application does not matter. This in turn implies that our likelihood $P(X|T)$ is not a mixture, which might be advantageous for computational reasons. Especially in Bayesian settings where computational concerns are prominent, being able to guarantee that the posterior is not a mixture is a desirable property to have.

\subsection{Sources of Noise in the Likelihood}
In many parts of this section, we assumed that the data $X$ did not lie exactly on the manifold $M$, and we often invoked a noisy data-generating process so that the likelihood $P(X|Z)$ did not collapse into a single point. While our results can be applied even if the likelihood collapses to a single point, the non-point likelihoods were more illustrative. Furthermore, assuming a noisy setting is not the only way to prevent the likelihood from collapsing. For example, $Z$ might be an partial set of coordinates so we are not able to specify the manifold location given $Z$. In that case, we may consider a noiseless setting and still have a non-point likelihood.

\section{Conclusion}
In this paper, we define disentanglement and factors of variation from the perspective of manifolds, unifying disentanglement with the manifold hypothesis. Under our framework, disentanglement amounts to recovering the data manifold's local coordinates. Along the way, we established that commutativity between factors is of central importance to disentanglement, capturing the property of separateness between factors. We studied the application of our framework to the problem of learning matrix exponentials, showing how enforcing commutativity makes the solution computationally tractable and easy to interpret. We then tackled the problem of distilling data-generating models, establishing the existence of a local procedure that simultaneously distills and disentangles a data-generating model. Finally, we related our manifold framework to other disentanglement frameworks: group theoretic and probabilistic. In each case, we demonstrated how each framework can be integrated within the manifold framework. Importantly, we also established an analogous commutativity criterion for each alternative framework, demonstrating that commutativity is central to disentanglement regardless of what framework one works in. We hope that this paper highlights the utility of the manifold framework of disentanglement and stimulates further research into the role of commutativity in disentanglement.

\section{Acknowledgements}
The author would like to thank Giles Hooker for his many helpful comments and suggestions in presenting the ideas of this paper. The author is also indebted to Bruno Olshausen, Yubei Chen, and Ho Yin Chau for the many stimulating discussion about disentanglement, and our previous joint work sparked the ideas contained  here.

\clearpage 
\bibliography{refs.bib}

\begin{thebibliography}{28}
\providecommand{\natexlab}[1]{#1}
\providecommand{\url}[1]{\texttt{#1}}
\expandafter\ifx\csname urlstyle\endcsname\relax
  \providecommand{\doi}[1]{doi: #1}\else
  \providecommand{\doi}{doi: \begingroup \urlstyle{rm}\Url}\fi

\bibitem[Bengio(2013)]{bengio2013deep}
Y.~Bengio.
\newblock Deep learning of representations: Looking forward, 2013.

\bibitem[Bengio et~al.(2014)Bengio, Courville, and
  Vincent]{bengio2014representation}
Y.~Bengio, A.~Courville, and P.~Vincent.
\newblock Representation learning: A review and new perspectives, 2014.

\bibitem[Brahma et~al.(2016)Brahma, Wu, and She]{DeepLearnWorksManiHyp}
P.~P. Brahma, D.~Wu, and Y.~She.
\newblock Why deep learning works: A manifold disentanglement perspective.
\newblock \emph{IEEE Transactions on Neural Networks and Learning Systems},
  27\penalty0 (10):\penalty0 1997--2008, 2016.
\newblock \doi{10.1109/TNNLS.2015.2496947}.

\bibitem[Burgess et~al.(2018)Burgess, Higgins, Pal, Matthey, Watters,
  Desjardins, and Lerchner]{burgess2018understanding}
C.~P. Burgess, I.~Higgins, A.~Pal, L.~Matthey, N.~Watters, G.~Desjardins, and
  A.~Lerchner.
\newblock Understanding disentangling in $\beta$-vae, 2018.

\bibitem[Chau et~al.(2020)Chau, Qiu, Chen, and
  Olshausen]{chau2020disentangling}
H.~Y. Chau, F.~Qiu, Y.~Chen, and B.~Olshausen.
\newblock Disentangling images with lie group transformations and sparse
  coding.
\newblock \emph{Arxiv}, 2020.

\bibitem[Chen et~al.(2018)Chen, Paiton, and Olshausen]{SparseManiTrans}
Y.~Chen, D.~Paiton, and B.~Olshausen.
\newblock The sparse manifold transform.
\newblock In S.~Bengio, H.~Wallach, H.~Larochelle, K.~Grauman, N.~Cesa-Bianchi,
  and R.~Garnett, editors, \emph{Advances in Neural Information Processing
  Systems}, volume~31. Curran Associates, Inc., 2018.

\bibitem[Cohen and Welling(2014)]{cohenLieGroup}
T.~Cohen and M.~Welling.
\newblock Learning the irreducible representations of commutative lie groups.
\newblock In \emph{Proceedings of the 31st International Conference on
  International Conference on Machine Learning - Volume 32}, ICML'14, page
  II–1755–II–1763. JMLR.org, 2014.

\bibitem[Fefferman et~al.(2016)Fefferman, Mitter, and
  Narayanan]{TestManiHypFefferman}
C.~Fefferman, S.~Mitter, and H.~Narayanan.
\newblock Testing the manifold hypothesis.
\newblock \emph{Journal of the American Mathematical Society}, 29:\penalty0
  983--1049, 2016.
\newblock \doi{10.1090/jams/852}.

\bibitem[Goyal and Bengio(2022)]{GoyalInducBias}
A.~Goyal and Y.~Bengio.
\newblock Inductive biases for deep learning of higher-level cognition.
\newblock \emph{Proceedings of the Royal Society A: Mathematical, Physical and
  Engineering Sciences}, 478\penalty0 (2266):\penalty0 20210068, 2022.
\newblock \doi{10.1098/rspa.2021.0068}.

\bibitem[Hall(2010)]{Hall_LieGroup}
B.~C. Hall.
\newblock \emph{{L}ie {G}roups, {L}ie {A}lgebras, and {R}epresentations: {A}n
  {E}lementary {I}ntroduction}, volume 222 of \emph{Graduate Texts in
  Mathematics}.
\newblock Springer, New York, 2010.
\newblock ISBN 9781441923134.

\bibitem[Higgins et~al.(2017)Higgins, Matthey, Pal, Burgess, Glorot, Botvinick,
  Mohamed, and Lerchner]{HigginsbetaVAE}
I.~Higgins, L.~Matthey, A.~Pal, C.~P. Burgess, X.~Glorot, M.~M. Botvinick,
  S.~Mohamed, and A.~Lerchner.
\newblock beta-vae: Learning basic visual concepts with a constrained
  variational framework.
\newblock In \emph{5th International Conference on Learning Representations,
  {ICLR} 2017, Toulon, France, April 24-26, 2017, Conference Track
  Proceedings}. OpenReview.net, 2017.

\bibitem[Higgins et~al.(2018)Higgins, Amos, Pfau, Racaniere, Matthey, Rezende,
  and Lerchner]{higgins2018definition}
I.~Higgins, D.~Amos, D.~Pfau, S.~Racaniere, L.~Matthey, D.~Rezende, and
  A.~Lerchner.
\newblock Towards a definition of disentangled representations.
\newblock \emph{Arxiv}, 2018.

\bibitem[Hyvärinen and Oja(2000)]{ICA_Hyvarinen}
A.~Hyvärinen and E.~Oja.
\newblock Independent component analysis: algorithms and applications.
\newblock \emph{Neural Networks}, 13\penalty0 (4):\penalty0 411--430, 2000.
\newblock ISSN 0893-6080.
\newblock \doi{https://doi.org/10.1016/S0893-6080(00)00026-5}.

\bibitem[Kim and Mnih(2018)]{KimDisByFact}
H.~Kim and A.~Mnih.
\newblock Disentangling by factorising.
\newblock In J.~Dy and A.~Krause, editors, \emph{Proceedings of the 35th
  International Conference on Machine Learning}, volume~80 of \emph{Proceedings
  of Machine Learning Research}, pages 2649--2658. PMLR, 10--15 Jul 2018.

\bibitem[Lee(2019)]{LeeRiemmanian}
J.~Lee.
\newblock \emph{Introduction to Riemannian Manifolds}.
\newblock Graduate Texts in Mathematics. Springer International Publishing,
  2019.
\newblock ISBN 9783319917542.

\bibitem[Lee(2003)]{Lee00}
J.~M. Lee.
\newblock \emph{Smooth Manifolds}.
\newblock Springer New York, New York, NY, 2003.
\newblock \doi{10.1007/978-0-387-21752-9_1}.

\bibitem[Locatello et~al.(2019)Locatello, Bauer, Lucic, Raetsch, Gelly,
  Sch{\"o}lkopf, and Bachem]{Locatello_disentanglement}
F.~Locatello, S.~Bauer, M.~Lucic, G.~Raetsch, S.~Gelly, B.~Sch{\"o}lkopf, and
  O.~Bachem.
\newblock Challenging common assumptions in the unsupervised learning of
  disentangled representations.
\newblock In K.~Chaudhuri and R.~Salakhutdinov, editors, \emph{Proceedings of
  the 36th International Conference on Machine Learning}, volume~97 of
  \emph{Proceedings of Machine Learning Research}, pages 4114--4124. PMLR,
  09--15 Jun 2019.

\bibitem[Miao and Rao(2007)]{Miao2007_learnLG}
X.~Miao and R.~P.~N. Rao.
\newblock Learning the lie groups of visual invariance.
\newblock \emph{Neural Comput.}, 19\penalty0 (10):\penalty0 2665--2693, Oct.
  2007.

\bibitem[Moler and Van~Loan(2003)]{MolerMatExpCost}
C.~Moler and C.~Van~Loan.
\newblock Nineteen dubious ways to compute the exponential of a matrix,
  twenty-five years later.
\newblock \emph{SIAM Review}, 45\penalty0 (1):\penalty0 3--49, 2003.
\newblock \doi{10.1137/S00361445024180}.

\bibitem[Reed et~al.(2014)Reed, Sohn, Zhang, and Lee]{LearnDisFactor}
S.~Reed, K.~Sohn, Y.~Zhang, and H.~Lee.
\newblock Learning to disentangle factors of variation with manifold
  interaction.
\newblock In E.~P. Xing and T.~Jebara, editors, \emph{Proceedings of the 31st
  International Conference on Machine Learning}, volume~32 of \emph{Proceedings
  of Machine Learning Research}, pages 1431--1439, Bejing, China, 22--24 Jun
  2014. PMLR.

\bibitem[Roweis and Saul(2000)]{LocallyLinearEmbed}
S.~T. Roweis and L.~K. Saul.
\newblock Nonlinear dimensionality reduction by locally linear embedding.
\newblock \emph{Science}, 290\penalty0 (5500):\penalty0 2323--2326, 2000.
\newblock \doi{10.1126/science.290.5500.2323}.

\bibitem[Sohl-Dickstein et~al.(2017)Sohl-Dickstein, Wang, and
  Olshausen]{sohldickstein2017unsupervised}
J.~Sohl-Dickstein, C.~M. Wang, and B.~A. Olshausen.
\newblock An unsupervised algorithm for learning lie group transformations,
  2017.

\bibitem[Tenenbaum et~al.(2000)Tenenbaum, de~Silva, and
  Langford]{Tenenbaum_isomap}
J.~B. Tenenbaum, V.~de~Silva, and J.~C. Langford.
\newblock A global geometric framework for nonlinear dimensionality reduction.
\newblock \emph{Science}, 290\penalty0 (5500):\penalty0 2319--2323, 2000.
\newblock \doi{10.1126/science.290.5500.2319}.

\bibitem[Wang et~al.(2021)Wang, Yue, Huang, Sun, and
  Zhang]{WangGroupDisentangle}
T.~Wang, Z.~Yue, J.~Huang, Q.~Sun, and H.~Zhang.
\newblock Self-supervised learning disentangled group representation as
  feature.
\newblock In M.~Ranzato, A.~Beygelzimer, Y.~Dauphin, P.~Liang, and J.~W.
  Vaughan, editors, \emph{Advances in Neural Information Processing Systems},
  volume~34, pages 18225--18240. Curran Associates, Inc., 2021.

\bibitem[Xiao and Liu(2020)]{xiaoContFlow}
C.~Xiao and L.~Liu.
\newblock Generative flows with matrix exponential.
\newblock In H.~D. III and A.~Singh, editors, \emph{Proceedings of the 37th
  International Conference on Machine Learning}, volume 119 of
  \emph{Proceedings of Machine Learning Research}, pages 10452--10461. PMLR,
  13--18 Jul 2020.

\bibitem[Xu and Ma(2012)]{XuLieGruopLearn}
Q.~Xu and D.~Ma.
\newblock Applications of lie groups and lie algebra to computer vision: A
  brief survey.
\newblock In \emph{2012 International Conference on Systems and Informatics
  (ICSAI2012)}, pages 2024--2029, 2012.
\newblock \doi{10.1109/ICSAI.2012.6223449}.

\bibitem[Zheng and Xue(2009)]{manifoldLearningChapter}
N.~Zheng and J.~Xue.
\newblock \emph{Manifold Learning}, pages 87--119.
\newblock Springer London, London, 2009.
\newblock ISBN 978-1-84882-312-9.
\newblock \doi{10.1007/978-1-84882-312-9_4}.

\bibitem[Zhou et~al.(2021)Zhou, Zelikman, Lu, Ng, Carlsson, and
  Ermon]{zhou2021evaluating}
S.~Zhou, E.~Zelikman, F.~Lu, A.~Y. Ng, G.~Carlsson, and S.~Ermon.
\newblock Evaluating the disentanglement of deep generative models through
  manifold topology, 2021.

\end{thebibliography}

\clearpage

\begin{appendices}
\section{Proofs}

\subsection*{Theorem \ref{Thm:FacDisRel}}
\begin{proof}
The first claim is a basic property of a local chart $(\phi, U)$, where the coordinate flows in $\phi(U)$ are mapped to smooth flows on $U$ via the diffeomorphism $\phi$. For the second claim, if the vector field of a local flow $\theta$ is 0 at $p$, then $\theta$ is constant at $p$ on some interval containing $0$ (Proposition 9.21 of \cite{Lee00}). Since our factors are non-degenerate, their vector fields must be non-zero at every point. The second claim then follows from Theorem 9.2.2 of \cite{Lee00}.
\end{proof}

\subsection*{Theorems \ref{Thm:CommCrit} and \ref{thm:flowVecEequiv}}
\begin{proof}
A set of smooth local flows commute if and only if their vectors fields commute by Theorem 9.44 of \cite{Lee00}, establishing Theorem \ref{thm:flowVecEequiv}. Since we assumed the vector fields are linearly independent, Theorem \ref{Thm:CommCrit} is a statement of the canonical form of commuting vector fields (Theorem 9.46 of \cite{Lee00}).
\end{proof}

\subsection*{Theorem \ref{thm:CommMatExp}}
\begin{proof}
The dictionary factors are global flows, since they are the action of the one-parameter subgroups $\gamma_i(t) = exp(t A_i)$ at each point $p$. Hence, the flows commute if and only if the one-parameter subgroups commute. By the Baker-Campbell-Hausdorff formula (Theorem 5.3 of \cite{Hall_LieGroup}), this is true if and only if the generators $A_i$ commute. It is well-known that commuting matrices can be simultaneously diagonalized, and every matrix can be diagonalized over $\mathbb{C}$.
\end{proof}

\subsection*{Theorem \ref{thm:DistDisenModel} and Corollary \ref{cor: DisMaxModels}}
\begin{proof}
    If the differential of a smooth function $f: \R^m \rightarrow M$ is full-rank at $x \in \R^m$, there exists an open $U \subseteq \R^m$ containing $x$ such that $f|_U$ is a smooth immersion or submersion (Proposition 4.1 of \cite{Lee00}). Hence, on this $U$ we may apply the Rank Theorem (Theorem 4.12 of \cite{Lee00}) to get Theorem \ref{thm:DistDisenModel}.

    The first claim of the corollary is a consequence of Theorem \ref{thm:DistDisenModel}. For the second claim, for every $p \in M$ there is an $x_p \in \R^m$ such that $f(x_p) = p$ since $f$ is surjective. Let $V_p$ and $U_p$ be the coordinate charts of Theorem \ref{thm:DistDisenModel} containg $x_p$ and $p$ respectively. The set $\set{U_p}_{p\in M}$ forms an open cover of $M$, so there exists a countable open subcover $\set{U_\alpha}_{\alpha \in \mathbb{Z}}$. Let $W_\alpha$ be the projection onto the first $n$-coordinates of the reparametrization $\psi(V_\alpha)$. We can then define the distilled version $f^*$ on the set $\bigcup_\alpha (W_\alpha \times \set{\alpha}) \subseteq \R^{n+1}$ as:
    \[
    f^*((w,\alpha)) = f(\psi_\alpha^{-1}{(w,0,\cdots,0)}) \qquad w \in W_\alpha, \ \alpha \in \mathbb{Z}
    \]
    Hence, using one more parameter to distinguish the local chart, $f^*$ is surjective and full-rank at every point in its latent space.
\end{proof}

\subsection*{Theorem \ref{Thm:GroupCommCrit}}
\begin{proof}
    Given the product group structure $G = G_1 \times \cdots \times G_n$, the subgroup associated with each factor must commute. As $f$ is a group homomorphism, their images $f(G_i)$ must also commute in $Aut(M)$. Conversely, suppose we have a collection of group homomorphisms $f_i : G_i \rightarrow Aut(M)$. Then, one can define a map $f: G \rightarrow Aut(M)$ via $f(g_1,\cdots,g_n) = f_1 (g_1) * \cdots * f_n(g_n)$. Since $f_i(G_i) \subseteq Aut(M)$ commute, $f$ is in fact a group homomorphism:
    \begin{align*}
        f(g_1*g'_1,\cdots,g_n g_n') &= f_1(g_1*g')*\cdots * f_n(g_n*g_n')\\
        &= f_1(g_1)*f_1(g'_1)* \cdots * f_n(g_n)*f_n(g_n')\\
        &= [f_1(g_1)*\cdots * f_n(g_n)] * [f_1(g_1')*\cdots * f_n(g_n')]\\
        &= f((g_1,\cdots,g_n)*(g_1',\cdots,g_n'))
    \end{align*}
    The second equality follows from the fact that each $f_i$ is a group homomorphism, and the third equality follows from commutativity of the subgroups $f_i(G_i) \subseteq Aut(M)$. The other properties of a group homomorphism are similarly checked using a similar computation.
\end{proof}

\subsection*{Theorem \ref{thm:GroupManiCorres}}
\begin{proof}
    For any $p \in M$, the orbit map $f_p: H \rightarrow M$ is an equivariant map, viewing $H$ as a $H$-set under self-multiplication. Hence, the orbit map is constant rank by Theorem 7.25 of \cite{Lee00}. Since $H$ is a product manifold with a product group action, we can apply the Rank Theorem to each factor group's coordinates. For $p \in M$, every $p' \in Orb(p)$ can be expressed as $p' = h' \cdot p $ for some $h' \in H$. Therefore, $f_{p'} = f_p \circ R_{h'}$ where $R_{h'}$ is right multiplication by $h'$. As $R_{h'}$ is a diffeomorphism, $f_{p'}$ has the same rank as $f_p$.
\end{proof}

\subsection*{Theorem \ref{thm:HomSpace}}
\begin{proof}
This is just a restatement of the Homogenous Space Characterization Theorem (Theorem 21.19 of \cite{Lee00}). The latter claim follows since $M$ is diffeomorphic to $H/Stab(p)$ under the above theorem and that $Stab(p) = e$ if $H$ acts freely.    
\end{proof}

\subsection*{Lemma \ref{Lem: CoordConsistent}}
\begin{proof}
    For the first claim, since $U$ is open and $M$ is equipped with the Borel $\sigma$-algebra, $\mu$ restricts to local measure $\mu|_U$. Equipping $\phi_i(U) \subseteq \R^n$ with its Borel $\sigma$-algebra inherited from $\R^n$, as $\phi_i$ is homeomorphism it is an invertible measurable map and the pushforward $\mu_i = \phi_i^*(\mu|_U)$ is unique. Since $U$ and $\phi_i(U)$ are related by an invertible measurable map, the only valid coordinate measure on $\phi_i(U)$ is $\phi_i^*(\mu|_U)$ in order to be consistent with $\mu|_U$. As usual, equality and uniqueness is up to sets of measure 0. To prove the second claim, we note that since $\phi_1$ is an invertible measurable map we have $\mu|_U = \phi^{-1*}_1 (\mu_1)$. We then calculate:
    \[
    \mu_2 = \phi_2^* (\mu|_U) = \phi_2 (\phi^{-1*}_1 (\mu_1)) = (\phi_2 \circ \phi_1^{-1})^* (\mu_1)
    \]
    As $\psi = \phi_2 \circ \phi_1^{-1}$, we have the second claim. To finish the third claim, we note that the transition map $\psi$ is also a homeomorphism and hence a invertible measurable map with respect to the Borel $\sigma$-algebras.
\end{proof}

\subsection*{Theorem \ref{thm:DistinguishLocalCoords}}
\begin{proof}
Let $Z$ and $\Tilde{Z}$ be distinct local coordinates of an open set $U \subseteq M$ with chart maps $\phi$ and $\Tilde{\phi}$ respectively. If they are distinct coordinate systems, there exists at least one point $x$ such that $\phi^{-1}(x) \neq \Tilde{\phi}^{-1}(x)$. Since $M$ is Hausdorff, there exist disjoint open sets $W$ and $\Tilde{W}$ containing $\phi^{-1}(x)$ and $\Tilde{\phi}^{-1}(x)$ respectively. Then, $V = \phi(W \cap U) \cap \Tilde{\phi}(\Tilde{W} \cap U)$ is an open set containing $x$ such that $\phi^{-1}(v) \neq \Tilde{\phi}^{-1}(v)$ for all $v \in V$. As $P(X|Z=v) = P(X|\phi^{-1}(v))$ and $P(X|\Tilde{Z} = v) = P(X|\Tilde{\phi}^{-1}(v))$, since the likelihoods are identifiable in $M$ we have the result. A similar proof establishes the result for a set of partial coordinates, as long as the family of likelihood are identifiable in $M$.
\end{proof}

\subsection*{Theorem \ref{thm:CommuCritProb}}
\begin{proof}
    In general, given the factor values $T$ the likelihood takes the form:
    \[
    P(X|T) = \sum_{\sigma \in S_k} P(\sigma) P(X|f_\sigma(T))
    \]
    where $f_\sigma(T)$ correspond to the action of the factors in order $\sigma$ and $P(\sigma)$ is our prior over the order. Since the family is identifiable and affinely independent, we see that the likelihood is not a mixture if and only if $P(X|f_\sigma(T))$ is the same for all $\sigma \in S_n$. This is equivalent to $f_\sigma$ being the same for all $\sigma$, which is equivalent to the factors commuting. 
\end{proof}

\section{Smooth Manifolds: Technical Overview}
We give a brief technical overview of smooth manifolds. For thorough references, we point the reader in the direction of Lee's introductory texts on smooth manifolds (\cite{Lee00}) and Riemmanian manifolds (\cite{LeeRiemmanian}) as well as Hall's text on Lie groups (\cite{Hall_LieGroup}).

\subsection{Smooth Manifolds}
A manifold is a space that locally looks like Euclidean space, much like how a sphere locally looks like a plane. More formally, a $n$-dimensional manifold $M$ is a second-countable Hausdorff space equipped with a collection of local charts $\phi_i: V_i \rightarrow U_i$. Here, the $V_i$ form an open cover of $M$, and each local chart is a homeomorphism between an open set in the manifold $V_i \subseteq M$ and an open set in Euclidean space $U_i \subseteq \R^n$. Unpacking this definition, a manifold is some space with a collection of small patches that tile it, and each patch is associated with its own chart map that gives a bijective mapping between the manifold patch and a corresponding patch of Euclidean $n$-space. If we consider each dimension of Euclidean space as a spatial coordinate, the chart map $\phi_i$ gives the $n$ local coordinates of the patch $V_i$. Hence, manifolds are locally equivalent to patches of Euclidean space.

A manifold is smooth if the transition maps between its local coordinates are smooth maps with smooth inverse. That is, consider the neighborhood $V_1 \cap V_2$, which is the intersection between the local charts $\phi_1: V_1 \rightarrow U_1$ and $\phi_2: V_2 \rightarrow U_2$. These give two coordinate systems describing $V_1 \cap V_2$, and we require that reparameterizing from coordinates $U_1$ to coordinates $U_2$ be a smooth function $\phi_1 \circ \phi_2^{-1}: U_1 \rightarrow U_2$ with smooth inverse. Note that the transition map is just a map between open subsets of Euclidean space, and so we can interpret smoothness in the usual Euclidean sense. In summary, a smooth manifold is a manifold where reparameterization of local coordinates is smooth.

Finally, we define smooth maps between smooth manifolds. We say a map $f: M \rightarrow N$ is smooth if at each $p \in M$ there exist open patches $U \subseteq M$ and $V \subseteq N$ such that: $p \in U$, $f(p) \in f(U) \subset V$, and the maps between their coordinates is smooth. That is, restricting the map $f$ to local coordinate patches, the map $f$ is a smooth map if it is smooth in local coordinates.

\subsection{Tangent Spaces, the Tangent Bundle, and Vector Fields}
Each point $p$ on the smooth manifold $M$ has an associated vector space called the tangent space,  denoted $T_p M$. Intuitively, the tangent space $T_p M$ is the set of directional derivatives along the manifold at $p$, and we can interpret an element of the tangent space $v \in T_p M$ as the derivative along velocity $v$. For example, consider the unit sphere. The tangent space at a point $p$ on the sphere is isomorphic as a vector space to $\R^2$, and visually it's represented by the tangent plane to the sphere at $p$. Every vector in this plane represents a possible direction we could move on that sphere. Note that the vector perpendicular to this plane does not lie in the tangent space - it represents a direction that takes us off the sphere.

Consider a smooth function between smooth manifolds $f: M \rightarrow N$.  At every $p \in M$, $f$ induces a linear map between the tangent spaces $T_p M$ and $T_{f(p)} N$, and we call this induced map the differential $\partial f_p$. Representing $f$ in local coordinates, the differential $\partial f_p$ coincides with the Jacobian of $f$ at $p$. Just as the Jacobian linearly maps directional derivatives in one space to another, the differential is a linear map between tangent spaces. As a linear map, we define the rank of the differential $\partial f_p$ to be its rank as a linear map. For a smooth map $f: M \rightarrow N$ with dim$(M) = m$ and dim$(N) = n$, we say that $f$ is a map of constant rank of its differential has the same rank at every $p \in M$. If $f$ has constant rank $m$ we say that it is a smooth immersion, and if $f$ has constant rank $n$ we say that it is a smooth submersion. Intuitively, smooth immersions are those that locally are smooth embeddings, where the coordinates of $M$ are a subset of those of $N$. Similarly, smooth submersions admit local sections, where the coordinates of $N$ are a subset of those of $M$.

Now we define vector fields over a manifold $M$. In the Euclidean case, a vector field assigns a directional derivative to each point, and analogously a vector field on a manifold $M$ assigns to each point $p \in M$ an element of its tangent space $T_p M$. If we take the disjoint union of the tangent spaces of each point in the manifold $\coprod T_p M$, the resulting set has a natural atlas of charts induced from $M$. This makes it into another manifold called the tangent bundle $TM$. Returning to vector fields, we define smooth vector fields as 
smooth functions $X: M \rightarrow TM$ such that $X(p) \in T_p M \subseteq TM$.

\subsection{Smooth Flows and Lie Derivatives}
We define a smooth flow on a manifold $M$ as a smooth function $\theta: D \times M \rightarrow M$, where $D \times p \subseteq \R$ is some open interval containing 0 for every $p$ such that:
\[
\theta(t,\theta(s,p)) = \theta(t+s. p) \qquad ; \qquad \theta(0,p) = p
\]
We may also restrict the flow to be defined locally over some open set $U \subset M$ rather than the entire manifold. 

Importantly the coordinates of a local chart $(U,\phi)$ have their associated flows. First, in coordinate space $\phi(U)$ consider the flow along the $i^{th}$ dimension:
\[
\Tilde{\theta}_i(t,x) = x + t e_i
\]
where $e_i$ is the $i^{th}$ coordinate vector. This flow has its corresponding vector field $\Tilde{X}_i$ that at each point $x$ assigns $\pdv{}{x_i}|_x \in T_x \R^n$. The flow and its associated vector field are related through the following relation
\[
\Tilde{X}_i(x) = \pard{}{t}|_{t=0} \Tilde{\theta}_i(t,x)
\]
and so we say that $\Tilde{X}_i$ is the infinitesimal generator of $\Tilde{\theta}_i$. In turn, through the inverse local chart map $\phi^{-1}$, $\Tilde{\theta}_i$ induce a flow $\theta_i$ on the manifold patch $U$. Similarly, the differential $\partial \phi^{-1}$ maps $\Tilde{X}_i$ to a corresponding vector field $X_i$ on $U$, and $X_i$ is the infinitesimal generator of $\theta_i$. We call $\theta_i$ the coordinate flows of the local chart, and every coordinate chart has its own unique set of coordinate flows. 

Intuitively, we can think of the infinitesimal generator $X$ of a smooth flow $\theta$ as defining dynamics over the manifold. If we started at $p \in M$ and followed the dynamics of $X$ for time $t$, we would end up at the point $\theta(t,p)$. In this way, the flow of a vector field returns the ending location on the manifold after starting at $p$ and following the vector field for time $t$.

Having defined the notion of a vector field and its associated smooth flow, we can now tackle the Lie derivative. The Lie derivative $L_{X_1} X_2$ of vector field $X_2$ with respect to vector field $X_1$ is another vector field whose value at every point is the change in $X_2$ along the flow of $X_1$. Point-wise it has the complicated form:
\[
L_{X_1} X_2|_p = \pdv{}{t}|_{t=0} \partial \theta_{-t}|_{\theta(t,p)} X_{\theta(t,p)}
\]
where $\theta_{-t}(p) = \theta(-t,p)$. Since the Lie derivative is antisymmetric, if $L_{X_1} X_2 = 0$ then  $L_{X_2} X_1 = 0$ also and vice versa: invariance under the other's flow is a symmetric property. It turns out that the Lie derivative coincides with the Lie bracket operation, a bilinear operation taking two vector fields and returning another vector field. The Lie bracket often gives an easier way to compute the Lie derivative between two vector fields: for example, in Euclidean coordinates the Lie bracket of the general linear group $GL(n,\R)$ is the matrix commutator bracket.

\end{appendices}

\end{document}